\documentclass[sigconf]{acmart}
\usepackage{threeparttable}
\usepackage{multirow}

\AtBeginDocument{%
  }

\setcopyright{cc}
\setcctype{by}

\copyrightyear{2026}
\acmYear{2026}
\acmDOI{10.1145/3774904.3792945}
\acmConference[WWW '26]{Proceedings of the ACM Web Conference 2026}{April 13--17, 2026}{Dubai, United Arab Emirates}
\acmBooktitle{Proceedings of the ACM Web Conference 2026 (WWW '26), April 13--17, 2026, Dubai, United Arab Emirates}
\begin{document}

\title[CitiLink-Summ: Summarization of Discussion Subjects in European Portuguese Municipal Meeting Minutes]{CitiLink-Summ: A Dataset of Discussion Subjects Summaries in European Portuguese Municipal Meeting Minutes}


\newcommand{\UPAffiliation}{%
  \additionalaffiliation{%
    \institution{Universidade do Porto}
    \city{Porto}
    \country{Portugal}
  }%
}
\newcommand{\UBIAffiliation}{%
  \additionalaffiliation{%
    \institution{University of Beira Interior}
    \city{Covilhã}
    \country{Portugal}
  }%
}

\author{Miguel Marques}
\orcid{0009-0002-7934-0173}
\authornotemark[1]
%\email{webmaster@marysville-ohio.com}
%\UBIAffiliation % tens de usar uma vez
\affiliation{%
    \institution{University of Beira Interior}
    \city{Covilhã}
    \country{Portugal}
  }%
\affiliation{%
  \institution{INESC TEC}
  \city{Porto}
  \country{Portugal}
}
\email{miguel.alexandre.marques@ubi.pt}

\author{Ana Luísa Fernandes}
%\UPAffiliation
\affiliation{%
    \institution{Universidade do Porto}
    \city{Porto}
    \country{Portugal}
  }%
\affiliation{%
   \institution{INESC TEC}
  \city{Porto}
  \country{Portugal}
}
\email{ana.l.fernandes@inesctec.pt}
\orcid{0009-0009-0552-3904}

\author{Ana Filipa Pacheco}
%\authornotemark[2]
\affiliation{%
    \institution{Universidade do Porto}
    \city{Porto}
    \country{Portugal}
  }%
\affiliation{%
   \institution{INESC TEC}
  \city{Porto}
  \country{Portugal}
}
\email{ana.f.pacheco@inesctec.pt}
\orcid{0009-0003-6135-8479}

\author{Rute Rebouças}
%\authornotemark[2]
\affiliation{%
    \institution{Universidade do Porto}
    \city{Porto}
    \country{Portugal}
  }%
\affiliation{%
   \institution{INESC TEC}
  \city{Porto}
  \country{Portugal}
}
\email{rute.reboucas@inesctec.pt}
\orcid{0000-0002-8213-1068}

\author{Inês Cantante}
%\authornotemark[2]
\affiliation{%
    \institution{Universidade do Porto}
    \city{Porto}
    \country{Portugal}
  }%
\affiliation{%
   \institution{INESC TEC}
  \city{Porto}
  \country{Portugal}
}
\email{ines.cantante@inesctec.pt}
\orcid{0009-0002-3866-4550}

\author{José Isidro}
%\authornotemark[2]
\affiliation{%
    \institution{Universidade do Porto}
    \city{Porto}
    \country{Portugal}
  }%
\affiliation{%
   \institution{INESC TEC}
  \city{Porto}
  \country{Portugal}
}
\email{jose.m.isidro@inesctec.pt}
\orcid{0009-0000-6071-9138}

\author{Luís Filipe Cunha}
%\authornotemark[2]
\affiliation{%
    \institution{Universidade do Porto}
    \city{Porto}
    \country{Portugal}
  }%
\affiliation{%
   \institution{INESC TEC}
  \city{Porto}
  \country{Portugal}
}
\email{luis.f.cunha@inesctec.pt}
\orcid{0000-0003-1365-0080}

\author{Alípio Jorge}
%\authornotemark[2]
\affiliation{%
    \institution{Universidade do Porto}
    \city{Porto}
    \country{Portugal}
  }%
\affiliation{%
   \institution{INESC TEC}
  \city{Porto}
  \country{Portugal}
}
\email{alipio.jorge@inesctec.pt}
\orcid{0000-0002-5475-1382}

\author{Nuno Guimarães}
\authornote{Corresponding author.}
\affiliation{%
    \institution{Universidade do Porto}
    \city{Porto}
    \country{Portugal}
  }%
%\authornotemark[2]
\affiliation{%
   \institution{INESC TEC}
  \city{Porto}
  \country{Portugal}
}
\email{nuno.r.guimaraes@inesctec.pt}
\orcid{0000-0003-2854-2891}

\author{Sérgio Nunes}
%\authornotemark[2]
\affiliation{%
    \institution{Universidade do Porto}
    \city{Porto}
    \country{Portugal}
  }%
\affiliation{%
   \institution{INESC TEC}
  \city{Porto}
  \country{Portugal}
}
\email{sergio.nunes@inesctec.pt}
\orcid{0000-0002-2693-988X}

\author{António Leal}
%\authornotemark[2]
\affiliation{%
    \institution{Universidade do Porto}
    \city{Porto}
    \country{Portugal}
  }%
\affiliation{%
  \institution{University of Macau}
  \city{Macau}
  \country{China}
}
\email{antonioleal@um.edu.mo}
\orcid{0000-0002-6198-2496}

\author{Purificação Silvano}
%\authornotemark[2]
\affiliation{%
    \institution{Universidade do Porto}
    \city{Porto}
    \country{Portugal}
  }%
\affiliation{%
   \institution{INESC TEC}
  \city{Porto}
  \country{Portugal}
}
\email{purificacao.silvano@inesctec.pt}
\orcid{0000-0001-8057-5338}

\author{Ricardo Campos}
\authornotemark[1]
\affiliation{%
    \institution{University of Beira Interior}
    \city{Covilhã}
    \country{Portugal}
  }%
\affiliation{
    \institution{INESC TEC}
    \city{Porto}
    \country{Portugal}
}
\email{ricardo.campos@ubi.pt}
\orcid{0000-0002-8767-8126}

\renewcommand{\shortauthors}{Miguel Marques et al.}

\begin{abstract}

Municipal meeting minutes are formal records documenting the discussions and decisions of local government, yet their content is often lengthy, dense, and difficult for citizens to navigate. Automatic summarization can help address this challenge by producing concise summaries for each discussion subject. Despite its potential, research on summarizing discussion subjects in municipal meeting minutes remains largely unexplored, especially in low-resource languages, where the inherent complexity of these documents adds further challenges. A major bottleneck is the scarcity of datasets containing high-quality, manually crafted summaries, which limits the development and evaluation of effective summarization models for this domain. In this paper, we present CitiLink-Summ, a new corpus of European Portuguese municipal meeting minutes, comprising 120 documents and 2,880 manually hand-written summaries, each corresponding to a distinct discussion subject. Leveraging this dataset, we establish baseline results for automatic summarization in this domain, employing state-of-the-art generative models (e.g., BART, PRIMERA) as well as large language models (LLMs), evaluated with both lexical and semantic metrics such as ROUGE, BLEU, METEOR, and BERTScore. CitiLink-Summ provides the first benchmark for municipal-domain summarization in European Portuguese, offering a valuable resource for advancing NLP research on complex administrative texts.

\end{abstract}

\begin{CCSXML}
<ccs2012>
   <concept>
       <concept_id>10002951.10003317.10003347.10003357</concept_id>
       <concept_desc>Information systems~Summarization</concept_desc>
       <concept_significance>500</concept_significance>
   </concept>
   <concept>
       <concept_id>10010147.10010178.10010179</concept_id>
       <concept_desc>Computing methodologies~Natural language processing</concept_desc>
       <concept_significance>300</concept_significance>
   </concept>
   <concept>
       <concept_id>10010147.10010178.10010179.10010186</concept_id>
       <concept_desc>Computing methodologies~Language resources</concept_desc>
       <concept_significance>300</concept_significance>
   </concept>
   <concept>
       <concept_id>10010405.10010476.10010936.10010938</concept_id>
       <concept_desc>Applied computing~E-government</concept_desc>
       <concept_significance>100</concept_significance>
   </concept>
</ccs2012>
\end{CCSXML}

\ccsdesc[500]{Information systems~Summarization}
\ccsdesc[500]{Computing methodologies~Natural language processing}
\ccsdesc[500]{Computing methodologies~Language resources}
%\ccsdesc[300]{Applied computing~E-government}

\keywords{Text Summarization, Municipal Meeting Minutes, European Portuguese, Natural Language Processing, Language Resources}



\settopmatter{printacmref=false} % Removes ACM reference format
\renewcommand\footnotetextcopyrightpermission[1]{} % removes footnote with conference information in first column

% Custom Attribution Statement for arXiv
\makeatletter
\def\@copyrightspace{\relax}
\makeatother

\thanks{© Authors 2026. This is the author's version of the work. It is posted here for your personal use. Not for redistribution. The definitive Version of Record was published in Proceedings of the ACM Web Conference 2026 (WWW '26), http://dx.doi.org/10.1145/3774904.3792945.}

\maketitle

\section{Introduction}

Municipalities produce large volumes of textual information through meeting minutes. Although these documents are essential for transparency and accountability, they are often lengthy, highly formal, and segmented into multiple discussion subjects, making them difficult for citizens to navigate. In this context, automatic text summarization (ATS) has become increasingly relevant for improving citizens' access to public information, enabling them to understand the main topics of the minutes without reading them in full. Despite progress in ATS, meeting summarization remains inherently challenging, as noted by \citet{Singh2021}: models must detect key topics, interpret the intent behind discussions, and filter out irrelevant information, all within a highly domain-specific context. These challenges are amplified in under-resourced languages such as European Portuguese, where both datasets and high-quality summaries in texts of administrative nature are scarce~\cite{almeida-amorim-2024-legal}, limiting the development of effective summarization methods in this domain. 

In this work, we present a new summarization corpus for European Portuguese, consisting of manually hand-written summaries of discussion subjects, from 120 municipal meeting minutes collected from six Portuguese Municipalities (Alandroal, Campo Maior, Covilhã, Fundão, Guimarães and Porto), covering the 2021-2024 administrative term. Each minute was first manually segmented into discussion subjects, and subsequently summarized by a team of four annotators with a linguistics background under the supervision of two expert linguists, following guidelines specifically developed for this project. The resulting corpus comprises 2,880 summaries, each corresponding to a discussion subject. The summarization process followed three phases (pre-writing, writing, and evaluation) to ensure the quality of the produced summaries. In addition, sensitive data was anonymized to preserve privacy. 

The main contributions of this work are threefold: 1) the creation of a new, domain-specific dataset in European Portuguese, comprising manually written summaries of discussion subjects from municipal meeting minutes; 2) baseline results using state-of-the-art encoder-decoder models (e.g., BART, PRIMERA) and LLMs (e.g., Gemini) evaluated with both lexical and semantic metrics; and 3) the release of all resources, including the corpus, the summarization guidelines, a dataset sample, and reproducibility code, publicly available on GitHub\footnote{\url{https://github.com/INESCTEC/citilink-summ}}.
\section{Related Work}

Automatic text summarization (ATS) follows the fundamental principles of summary writing, aiming to produce a shorter text that preserves the essential information of the original. Approaches to ATS are generally classified as extractive, which selects salient sentences from the original text, or abstractive, which generate new text while maintaining the main ideas~\cite{Giarelis2023}, with some studies adopting hybrid strategies~\cite{10421343}. While summarization has been widely applied in domains such as medicine, law, journalism, and scientific literature \cite{dahan-stanovsky-2025-state}, meeting-related documents represent a particularly challenging domain due to their length, structured nature, and the need to capture multiple discussion topics. This has motivated the creation of specialized datasets to support research on meeting summarization, including MeetingBank~\cite{Hu2023}, ICSI~\cite{Janin2003}, and AMI~\cite{AMI2006}. MeetingBank is built from city-council meeting transcripts and provides professionally edited abstractive summaries aligned to shorter meeting segments, totaling 6,892 segment-level summarization instances. ICSI contains transcripts from 75 meetings, each approximately one hour long, while AMI comprises 100 hours of recorded meetings with detailed annotations, including speaker turns and both extractive and abstractive summaries \cite{Carletta2006}. Beyond English, \citet{Shirafuji2020} studied summarization of Japanese assembly minutes, and \citet{nedoluzhko-etal-2022-elitr} introduced a Czech parliamentary meetings dataset with abstractive summaries aligned to entire sessions. These resources demonstrate the growing interest in meeting-related documents summarization and highlight the lack of comparable datasets in other languages and administrative contexts. While they focus primarily on transcripts, municipal meeting minutes differ in that they are lengthy, highly formal administrative documents that record the outcomes of complex meeting-subject discussions, mixing essential decisions with extended contextual material of varying relevance, rather than the interactions themselves.
To the best of our knowledge, Citilink-Summ is the first dataset of discussion-subject summaries derived from municipal meeting minutes in European Portuguese. It also introduces the first benchmark for summarization in this domain, providing a solid foundation for future research in this under-resourced language variety.

\section{The CitiLink-Summ Dataset}

\begin{figure*}
    \centering
    \includegraphics[width=0.8\linewidth]{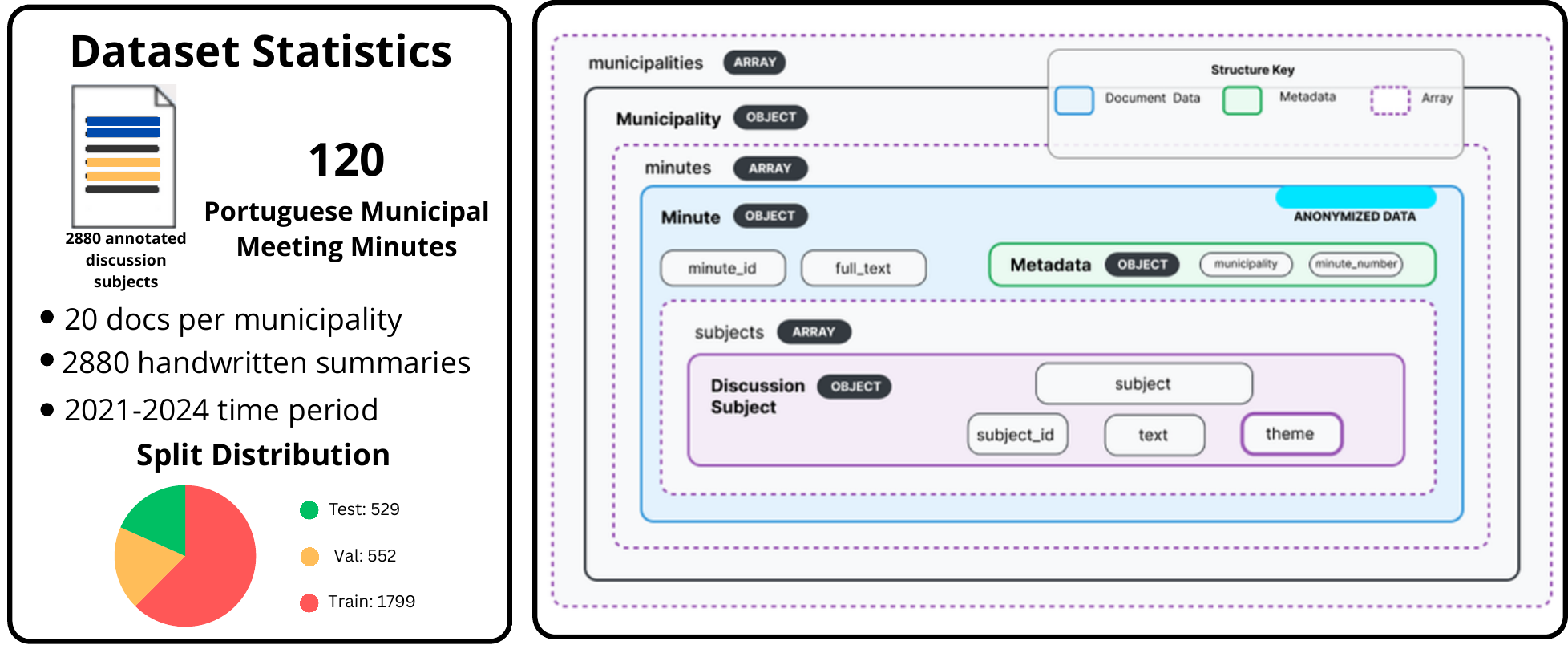}
    \caption{Dataset Statistics and JSON Schema.}
    \label{fig:overview}
\end{figure*}

% Still missing description of the final version in the text

The CitiLink-Summ dataset was built from Portuguese municipal meeting minutes and is designed to support segment-level summarization, with each segment corresponding to a distinct discussion subject. The dataset comprises 120 municipal meeting minutes written in European Portuguese, covering the 2021–2024 administrative term. It includes minutes from six municipalities: Alandroal, Campo Maior, Covilhã, Fundão, Guimarães and Porto, for which authorized access to official records was obtained. All minutes were collected, manually segmented into discussion subjects, and subsequently annotated with handwritten summaries. Sensitive information was manually de-identified to ensure privacy. In total, the dataset contains 2,880 discussion-subject segments, which were divided into training (60\%), validation (20\%) and test (20\%) sets using a random split within each municipality while preserving a proportional distribution across municipalities. An overview of dataset statistics and the JSON schema is shown in Figure~\ref{fig:overview}. The dataset follows a hierarchical structure beginning with a collection of municipalities, each containing multiple municipal meeting minutes. Every minute includes the full document text along with metadata, such as the municipality identifier and meeting number. Within each minute, a list of discussion subjects is provided, where each segment is represented by an object containing a unique identifier, the subject text and theme, as it makes part of a bigger annotation schema \cite{citilink2025}.

\begin{figure}[h!]
    \centering
    \includegraphics[width=\linewidth]{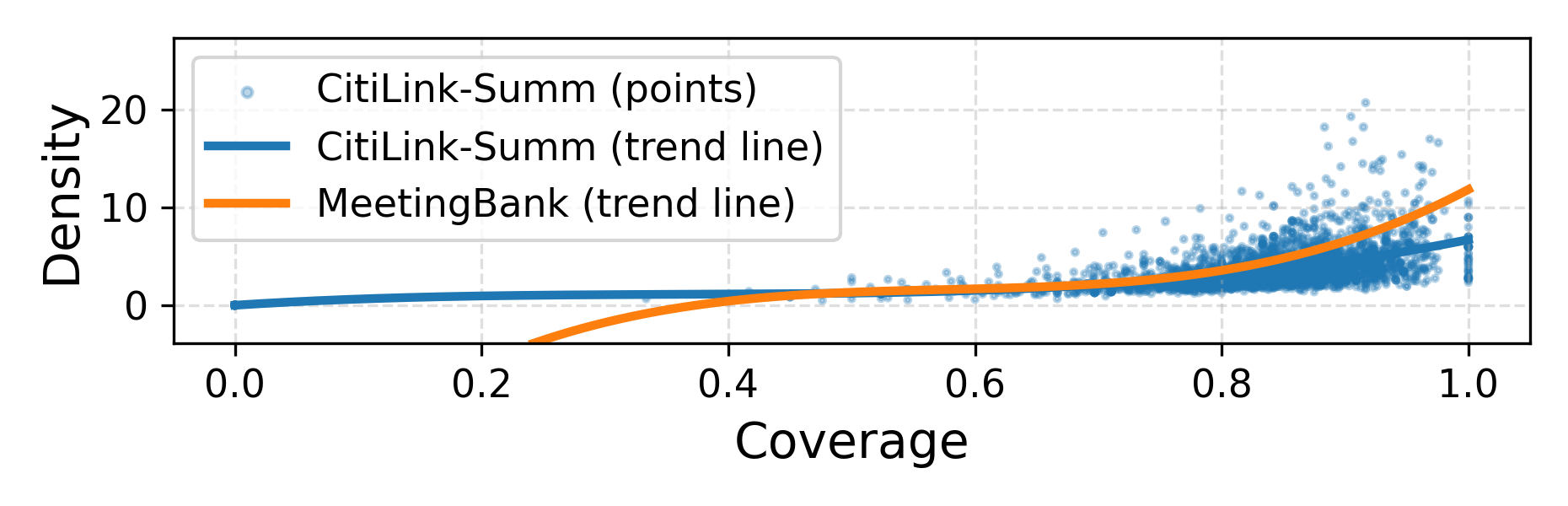}
    \caption{Overall relationship between coverage and density.}
    \label{fig:img0}
\end{figure}
To minimize subjectivity and ensure consistency across summaries, annotators followed strict guidelines and textual templates. The summarization workflow adopted builds on established practices \cite{friend2002summing} and consists of several steps, which we briefly outline below. 

Four annotators, PhD students in Linguistics with substantial annotation experience, supervised by two expert linguists, were responsible for manually handwriting the summaries\footnote{The full summarization guidelines are available on our GitHub repository.}. For each discussion-subject segment, annotators first read the full segment carefully, re-reading whenever necessary to ensure complete understanding, and then defined the theme of the discussion subject. Next, they identified the segment's main idea and expressed it concisely, preferably in their own words, while ensuring that the summary preserved the essential information without becoming vague or overly detailed. A draft summary was produced and subsequently revised by the annotator. In the final stage, annotators evaluated each summary according to five predefined parameters, including relevance, cohesion, coherence, and length. Summaries that did not achieve the maximum score in any parameter were revised and rewritten, returning to the previous stage. Once a summary reached the highest score across all parameters, it was deemed complete. This careful multi-stage annotation process not only ensured high-quality summaries but also intentionally shaped their level of abstraction, a key aspect given that higher abstraction generally increases the difficulty for models to generate accurate outputs. By enforcing multiple revisions, strict quality controls, and clear guidelines, annotators were encouraged to reformulate and condense information rather than simply extract text, thereby mitigating the challenges associated with high abstraction.

To quantify this effect, we analyzed the level of abstraction in the produced summaries, providing insights into the linguistic transformations required to generate high-quality summaries and highlighting the challenges posed by the dataset. Two commonly used metrics quantify abstraction by measuring reused text: Coverage and Density \cite{grusky-etal-2018-newsroom}. Coverage calculates the fraction of summary tokens that appear in the source text, while Density captures the extent to which a summary consists of contiguous extractive fragments \cite{Hu2023}. As shown in Figure~\ref{fig:img0}, CitiLink-Summ exhibits low Density and medium-to-high Coverage, similar to other meeting-related datasets such as MeetingBank, whose trend line is shown in orange. This indicates that while summaries reuse vocabulary from the source, they rarely rely on long extractive spans, remaining largely abstractive and requiring substantial reformulation to preserve key information.

\section{Benchmark Evaluation}

\begin{table*}[h!]
\setlength{\belowcaptionskip}{3pt}
\centering
\caption{Performance of summarization models on the CitiLink-Summ dataset using lexical and semantic metrics.}
\label{tab:summary_comparison}
\begin{threeparttable}
\resizebox{0.8\linewidth}{!}{
\setlength{\tabcolsep}{4pt}
\renewcommand{\arraystretch}{0.8}
\begin{tabular}{
    l
    |ccc
    |cc
    |ccc
}
\toprule
\textbf{Model} & \multicolumn{3}{c|}{\textbf{ROUGE}} & \multicolumn{2}{c|}{\textbf{BLEU + METEOR}} & \multicolumn{3}{c}{\textbf{BERTSCORE}} \\
 & R-1 & R-2 & R-L & BLEU & METEOR & F1 & PREC & RECALL  \\
\midrule
BART        & 63.52 & 49.22 & 58.20 & 36.06 & 55.15 & 83.87 & 84.59 & 83.37  \\ %updated
BART Large  & 68.96 & 54.78 & 63.64 & 42.43 & 61.65 & 86.28 & 86.57 & 86.15  \\ %updated
PTT5        & 52.36 & 38.21 & 45.26 & 23.65 & 46.44 & 76.90 & 76.12 & 78.15  \\ %updated
LED         & 63.63 & 50.50 & 58.59 & 29.82 & 54.88 & 84.16 & 85.70 & 82.91  \\ %updated
PRIMERA     & 66.17 & 54.57 & 61.94 & 29.06 & 57.05 & 85.79 & 87.10 & 84.79  \\
\midrule
Qwen2.5-1.5B        & 44.24 & 31.06 & 38.80 & 07.16 & 31.79 & 74.49 & 77.75 & 71.83  \\ %updated
Gemini-2.5-flash      & 64.16 & 48.94 & 55.97 & 28.40 & 54.34 & 83.09 & 82.99 & 83.19  \\ %updated
\bottomrule
\end{tabular}
}
\end{threeparttable}
\end{table*}

To establish reference benchmarks for CitiLink-Summ, we evaluated several state-of-the-art summarization models, fine-tuning them on our dataset to obtain comparative results. The summarization pipeline comprises three main stages: (i)~chunking; (ii)~model selection and fine-tuning; and (iii)~evaluation of the generated summaries. In the first stage, lightweight chunking was applied. Due to the limited context window of the language models (256-2048 tokens), each discussion-subject was divided hierarchically using a sliding-window chunking approach. Each discussion subject was split into smaller chunks, summarized individually, and aggregated to form the final segment-level summary. 

Formally, each instance in CitiLink-Summ consists of a discussion subject $s$, and its associated theme $\tau$. While the manually written summaries were produced with knowledge of the theme, the benchmark models are evaluated using only the discussion-subject text. It is important to note that, although this represents a conceptual limitation compared to the annotation process, it reflects the practical goal of generating summaries without dependencies beyond the text itself. Consequently, the reported results represent a lower-bound performance, and do not fully capture potential improvements achievable if models had access to the additional contextual information provided to the annotators.

For benchmark evaluation, we selected pre-trained encoder-decoder models commonly used in abstractive summarization, including BART and BART Large~\cite{lewis2019bart}, PTT5~\cite{2020arXiv200809144C}, LED~\cite{beltagy2020longformer}, and PRI\-MERA~\cite{xiao2022primera}, which were fine-tuned to adapt to the terminology of our corpus, and are expected to internalize additional data-specific patterns that are not explicitly present in the reference summaries. In addition, we evaluated two large generative models, Qwen2.5-1.5B-Instruct (open source) and Gemini-2.5-flash (closed source), using few-shot prompting, to compare the performance of traditional fine-tuned summarizers with modern foundation models. The models were evaluated using several metrics known to be reliable quality indicators for summarization tasks ~\cite{fabbri-etal-2021-summeval}. Particularly, to capture model performance comprehensively, we employed both lexical and semantic metrics. Lexical metrics, such as ROUGE~\cite{lin2004rouge}, BLEU~\cite{papineni2002bleu}, and METEOR~\cite{banerjee2005meteor}, measure token and n-gram overlap with the reference summaries, with METEOR additionally accounting for stemming and synonymy to better align linguistically. Semantic metrics, such as BERTScore~\cite{zhang2020bertscore}, evaluate contextual similarities, complementing lexical metrics by assessing semantic preservation that purely word-level comparisons may miss.

Table~\ref{tab:summary_comparison} reports the results of all selected models and evaluation metrics on CitiLink-Summ, providing a benchmark for future research on European Portuguese municipal meeting summarization.
%\subsection{Results Analysis}
The results demonstrate that the chosen models are capable of producing abstractive summaries with coherent information, although their performance indicates that there is still considerable room for improvement. As expected, larger models, such as PRIMERA, BART large, and Gemini, achieved the highest scores across all metrics. %The PRIMERA model is the top performer on the task, given that it achieved the best scores across most metrics. %In addition, Primera not only achieves the highest on quantitative metrics, but, with a result of XXX on X,Y,Z metrics, also produces higher-quality, more readable, and more grammatically well-formed summaries.
Although the absolute scores remain moderate, the models exhibit consistent behavior across metrics, suggesting that they can serve as a solid foundation for future work aimed at improving summarization quality for municipal meeting minutes.

\section{Conclusions}

In this paper, we introduce the first dataset benchmark for the abstractive text summarization of discussion subjects from municipal meeting minutes in European Portuguese, addressing the lack of resources for this under-explored task. In addition, we provide baseline models for this task, with the evaluation demonstrating the potential of current summarization techniques in this challenging context. Notably, larger models like PRIMERA achieved the best overall performance across most metrics, showing that state-of-the-art approaches can be effectively applied to municipal meeting minutes. These contributions establish a foundation for research on Portuguese municipal-domain summarization. Future work will focus on expanding the corpus, incorporating human evaluations to ensure the generated summaries meet standards of clarity and accuracy, and exploring approaches that leverage theme information to further improve performance. Additionally, we plan to release an English version of the dataset to increase accessibility and enable cross-lingual research. The dataset is publicly released to support the development of more advanced models and foster research in this under-resourced language. Ultimately, this resource is expected to facilitate the creation of practical summarization systems, improving the accessibility and transparency of municipal decision-making for the general public.

\begin{acks}
This work is funded by national funds through FCT – Fundação para a Ciência e a Tecnologia, I.P., under the support UID/50014/2025 (\url{https://doi.org/10.54499/UID/50014/2025}).
The authors would also like to acknowledge the project CitiLink, with reference 2024.07509 IACDC (\url{https://doi.org/10.54499/2024.07509.IACDC}).
\end{acks}

\bibliographystyle{ACM-Reference-Format}
\bibliography{biblio}

@ARTICLE{Giarelis2023,
  author       = {Nikolaos Giarelis and Charalampos Mastrokostas and Nikos Karacapilidis},
  title        = {Abstractive vs. Extractive Summarization: An Experimental Review},
  journal      = {Applied Sciences},
  volume       = {13},
  number       = {13},
  pages        = {7620},
  year         = 2023,
  doi          = {10.3390/app13137620}
}

@inproceedings{Hu2023,
  author       = {Yebowen Hu and Timothy Ganter and Hanieh Deilamsalehy and Franck Dernoncourt and Hassan Foroosh and Fei Liu},
  title        = {MeetingBank: A Benchmark Dataset for Meeting Summarization},
  booktitle    = {Proceedings of the 61st Annual Meeting of the Association for Computational Linguistics (ACL)},
  year         = 2023,
  month        = {July},
  address      = {Toronto, Canada},
  publisher    = {Association for Computational Linguistics},
  url          = {https://aclanthology.org/2023.acl-long.906/}
}

@ArtifactDataset{AMI2006,
  author       = {Jean Carletta and Mark Ashby and Jérémy Boudy and John Garofalo and Dafydd Gibbon and Stefan Götz and Thomas Hain and Jun He and John Hough and Bernhard Krenn and Lori Lamel and Johanna Moore and David O'Neill and Conor O'Riordan and Steve Renals and Shane Rickard and Peter Robinson and Stephanie Seneff and Paul Taylor},
  title        = {The AMI Meeting Corpus},
  year         = 2006,
  url          = {https://groups.inf.ed.ac.uk/ami/corpus/},
  lastaccessed = {September 17, 2025}
}

@ArtifactDataset{Janin2003,
  author       = {Adam Janin and Don Baron and Jane Edwards and Dan Ellis and David Gelbart and Nelson Morgan and Barbara Peskin and Thilo Pfau and Elizabeth Shriberg and Andreas Stolcke and Chuck Wooters},
  title        = {The ICSI Meeting Corpus},
  year         = 2003,
  url          = {https://www.ee.columbia.edu/~dpwe/pubs/icassp03-janin.pdf},
  lastaccessed = {September 17, 2025}
}

@Preprint{Shirafuji2020,
  author       = {Daiki Shirafuji and Hiromichi Kameya and Rafal Rzepka and Kenji Araki},
  title        = {Summarizing Utterances from Japanese Assembly Minutes using Political Sentence-BERT-based Method for QA Lab-PoliInfo-2 Task of NTCIR-15},
  year         = 2020,
  url          = {https://arxiv.org/abs/2010.12077},
  doi          = {10.48550/arXiv.2010.12077}
}

@inproceedings{Singh2021,
  author       = {Manpreet Singh and Tirthankar Ghosal and Ond{\v{r}}ej Bojar},
  title        = {An Empirical Analysis of Text Summarization Approaches for Automatic Minuting},
  booktitle    = {Proceedings of the 35th Pacific Asia Conference on Language, Information and Computation},
  year         = 2021
}

@ARTICLE{Carletta2006,
  author       = {J. Carletta},
  title        = {Announcing the AMI Meeting Corpus},
  journal      = {The ELRA Newsletter},
  volume       = {11},
  number       = {1},
  pages        = {3--5},
  month        = {January--March},
  year         = 2006
}

@inproceedings{grusky-etal-2018-newsroom,
    title = "{N}ewsroom: A Dataset of 1.3 Million Summaries with Diverse Extractive Strategies",
    author = "Grusky, Max  and
      Naaman, Mor  and
      Artzi, Yoav",
    editor = "Walker, Marilyn  and
      Ji, Heng  and
      Stent, Amanda",
    booktitle = "Proceedings of the 2018 Conference of the North {A}merican Chapter of the Association for Computational Linguistics: Human Language Technologies, Volume 1 (Long Papers)",
    month = jun,
    year = "2018",
    address = "New Orleans, Louisiana",
    publisher = "Association for Computational Linguistics",
    url = "https://aclanthology.org/N18-1065/",
    doi = "10.18653/v1/N18-1065",
    pages = "708--719"
}

@article{lewis2019bart,
  title={BART: Denoising sequence-to-sequence pre-training for natural language generation, translation, and comprehension},
  author={Lewis, Mike and Liu, Yinhan and Goyal, Naman and Ghazvininejad, Marjan and Mohamed, Abdelrahman and Levy, Omer and Stoyanov, Veselin and Zettlemoyer, Luke},
  journal={arXiv preprint arXiv:1910.13461},
  year={2019}
}

@article{beltagy2020longformer,
  title={Longformer: The long-document transformer},
  author={Beltagy, Iz and Peters, Matthew E. and Cohan, Arman},
  journal={arXiv preprint arXiv:2004.05150},
  year={2020}
}

@inproceedings{xiao2022primera,
  title={PRIMERA: Pyramid-based masked sentence pre-training for multi-document summarization},
  author={Xiao, Wen and Carenini, Giuseppe},
  booktitle={Proceedings of the 60th Annual Meeting of the Association for Computational Linguistics (ACL)},
  year={2022}
}

@inproceedings{lin2004rouge,
  title={ROUGE: A package for automatic evaluation of summaries},
  author={Lin, Chin-Yew},
  booktitle={Text Summarization Branches Out: Proceedings of the ACL-04 Workshop},
  year={2004},
  pages={74--81}
}

@inproceedings{papineni2002bleu,
  title={BLEU: a method for automatic evaluation of machine translation},
  author={Papineni, Kishore and Roukos, Salim and Ward, Todd and Zhu, Wei-Jing},
  booktitle={Proceedings of the 40th Annual Meeting of the Association for Computational Linguistics},
  pages={311--318},
  year={2002}
}

@inproceedings{banerjee2005meteor,
  title={METEOR: An automatic metric for MT evaluation with improved correlation with human judgments},
  author={Banerjee, Satanjeev and Lavie, Alon},
  booktitle={Proceedings of the ACL Workshop on Intrinsic and Extrinsic Evaluation Measures for Machine Translation and/or Summarization},
  pages={65--72},
  year={2005}
}

@inproceedings{zhang2020bertscore,
  title={BERTScore: Evaluating text generation with BERT},
  author={Zhang, Tianyi and Kishore, Varsha and Wu, Felix and Weinberger, Kilian Q. and Artzi, Yoav},
  booktitle={International Conference on Learning Representations (ICLR)},
  year={2020}
}

@inproceedings{almeida-amorim-2024-legal,
    title = "A Legal Framework for Natural Language Model Training in {P}ortugal",
    author = "Almeida, Ruben  and
      Amorim, Evelin",
    editor = "Siegert, Ingo  and
      Choukri, Khalid",
    booktitle = "Proceedings of the Workshop on Legal and Ethical Issues in Human Language Technologies @ LREC-COLING 2024",
    month = may,
    year = "2024",
    address = "Torino, Italia",
    publisher = "ELRA and ICCL",
    url = "https://aclanthology.org/2024.legal-1.2/",
    pages = "6--12"
}

@inproceedings{dahan-stanovsky-2025-state,
    title = "The State and Fate of Summarization Datasets: A Survey",
    author = "Dahan, Noam  and
      Stanovsky, Gabriel",
    editor = "Chiruzzo, Luis  and
      Ritter, Alan  and
      Wang, Lu",
    booktitle = "Proceedings of the 2025 Conference of the Nations of the Americas Chapter of the Association for Computational Linguistics: Human Language Technologies (Volume 1: Long Papers)",
    month = apr,
    year = "2025",
    address = "Albuquerque, New Mexico",
    publisher = "Association for Computational Linguistics",
    url = "https://aclanthology.org/2025.naacl-long.372/",
    doi = "10.18653/v1/2025.naacl-long.372",
    pages = "7259--7278",
    ISBN = "979-8-89176-189-6"
}

@article{fabbri-etal-2021-summeval,
    title = "{S}umm{E}val: Re-evaluating Summarization Evaluation",
    author = "Fabbri, Alexander R.  and
      Kry{\'s}ci{\'n}ski, Wojciech  and
      McCann, Bryan  and
      Xiong, Caiming  and
      Socher, Richard  and
      Radev, Dragomir",
    editor = "Roark, Brian  and
      Nenkova, Ani",
    journal = "Transactions of the Association for Computational Linguistics",
    volume = "9",
    year = "2021",
    address = "Cambridge, MA",
    publisher = "MIT Press",
    url = "https://aclanthology.org/2021.tacl-1.24/",
    doi = "10.1162/tacl_a_00373",
    pages = "391--409"
}

@inproceedings{nedoluzhko-etal-2022-elitr,
    title = "{ELITR} Minuting Corpus: A Novel Dataset for Automatic Minuting from Multi-Party Meetings in {E}nglish and {C}zech",
    author = "Nedoluzhko, Anna  and
      Singh, Muskaan  and
      Hled{\'i}kov{\'a}, Marie  and
      Ghosal, Tirthankar  and
      Bojar, Ond{\v{r}}ej",

    booktitle = "Proceedings of the Thirteenth Language Resources and Evaluation Conference",
    month = jun,
    year = "2022",
    address = "Marseille, France",
    publisher = "European Language Resources Association",
    url = "https://aclanthology.org/2022.lrec-1.340/",
    pages = "3174--3182"
}

@INPROCEEDINGS{10421343,
  author={Rakshitha and Mohan, Pushpa and Shanthi, M B and Disha, D N and Rao, Sudesh},
  booktitle={2023 International Conference on Integrated Intelligence and Communication Systems (ICIICS)}, 
  title={Advances in Natural Language Processing and Deep Learning for Document Summarization}, 
  year={2023},
  volume={},
  number={},
  pages={1-6},
  doi={10.1109/ICIICS59993.2023.10421343}}

@article{friend2002summing,
  title={Summing it up},
  author={Friend, Rosalie},
  journal={The Science Teacher},
  volume={69},
  number={4},
  pages={40},
  year={2002},
  publisher={Taylor \& Francis Ltd.}
}

@ARTICLE{2020arXiv200809144C,
       author = {{Carmo}, Diedre and {Piau}, Marcos and {Campiotti}, Israel and {Nogueira}, Rodrigo and {Lotufo}, Roberto},
        title = "{PTT5: Pretraining and validating the T5 model on Brazilian Portuguese data}",
      journal = {arXiv e-prints},
         year = 2020,
        month = aug,
          eid = {arXiv:2008.09144},
        pages = {arXiv:2008.09144},
          doi = {10.48550/arXiv.2008.09144},
archivePrefix = {arXiv},
       eprint = {2008.09144},
 primaryClass = {cs.CL},
       adsurl = {https://ui.adsabs.harvard.edu/abs/2020arXiv200809144C}
}

@dataset{citilink2025,
  author       = {Ricardo Campos and Ana Filipa Pacheco and Ana Luísa Fernandes and Inês Cantante and Rute Rebouças and Luís Filipe Cunha and José Isidro and José Evans and Miguel Marques and Rodrigo Batista and Evelin Amorim and Alípio Jorge and Nuno Guimarães and Sérgio Nunes and António Leal and Purificação Silvano},
  title        = {CitiLink-Minutes: A Multilayer Annotated Dataset of Municipal Meeting Minutes},
  year         = {2025},
  doi          = {https://doi.org/10.25747/7KG6-1K22},
  institution  = {INESC TEC}
}

\end{document}